%% file: main.tex
\documentclass[10pt,twocolumn,letterpaper]{article}

\usepackage[pagenumbers]{cvpr} 

\input{preamble}

\definecolor{cvprblue}{rgb}{0.21,0.49,0.74}
\usepackage[pagebackref,breaklinks,colorlinks,citecolor=cvprblue]{hyperref}

\def\paperID{3233} 
\def\confName{CVPR}
\def\confYear{2024}

\title{Generating Content for HDR Deghosting from Frequency View}

\author{Tao Hu$^{1\dag}$ \ \ \ \ Qingsen Yan$^{1\dag}$ \thanks{Corresponding author. $\dag$~The first two authors contributed equally to this work.
This work was partially supported by NSFC (62301432,62306240), NSBRPS (2023-JC-QN-0685, QCYRCXM-2023-057). Y. Qi is not supported by the above mentioned fundings. 
}\ \ \ \ Yuankai Qi$^{2}$ \ \ \ \ Yanning Zhang$^{1}$\\
$^{1}$Northwestern Polytechnical University\ $^{2}$Macquarie University\\
{\tt\small hutaoxauat@gmail.com,$^{1}$\{qingsenyan,ynzhang\}@nwpu.edu.cn,$^{2}$yuankai.qi@mq.edu.au}
}

\begin{document}
\maketitle
\input{sec/abstrct}
\input{sec/1_intro}
\input{sec/2_methods}

\input{sec/3_experiments}
{
    \small
    \bibliographystyle{ieeenat_fullname}
    \bibliography{main}
}


\end{document}

%% file: preamble.tex
\usepackage[dvipsnames]{xcolor}
\usepackage{tabularx, booktabs}
\newcommand{\red}[1]{{\color{red}#1}}

\usepackage{tabularx}
\usepackage{booktabs}
\usepackage{adjustbox}
\usepackage{tabularx}
\usepackage{booktabs}
\usepackage{graphicx}
\usepackage{multirow}
\usepackage{indentfirst}
\usepackage{pifont}
\usepackage{algorithm}  
\usepackage{algorithmic} 

%% file: sec/abstrct.tex
\begin{abstract}
Recovering ghost-free High Dynamic Range (HDR) images from multiple Low Dynamic Range (LDR) images becomes challenging when the LDR images exhibit saturation and significant motion. 
Recent Diffusion Models (DMs) have been introduced in HDR imaging field, demonstrating promising performance, particularly in achieving visually perceptible results compared to previous DNN-based methods. 
However, DMs require extensive iterations with large models to estimate entire images, resulting in inefficiency that hinders their practical application.
To address this challenge, we propose the Low-Frequency aware Diffusion (LF-Diff) model for ghost-free HDR imaging. 
The key idea of LF-Diff is implementing the DMs in a highly compacted latent space and integrating it into a regression-based model to enhance the details of reconstructed images. 
Specifically, as low-frequency information is closely related to human visual perception we propose to utilize DMs to create compact low-frequency priors for the reconstruction process. 
In addition, to take full advantage of the above low-frequency priors, the Dynamic HDR Reconstruction Network (DHRNet) is carried out in a regression-based manner to obtain final HDR images.
Extensive experiments conducted on synthetic and real-world benchmark datasets demonstrate that our LF-Diff performs favorably against several state-of-the-art methods and is 10$\times$ faster than previous DM-based methods.
\end{abstract}


%% file: sec/1_intro.tex
\def \red{\textcolor{red}}

\section{Introduction}
\label{sec:intro}
\begin{figure}[tb]
\centering
\includegraphics[width=1\linewidth]{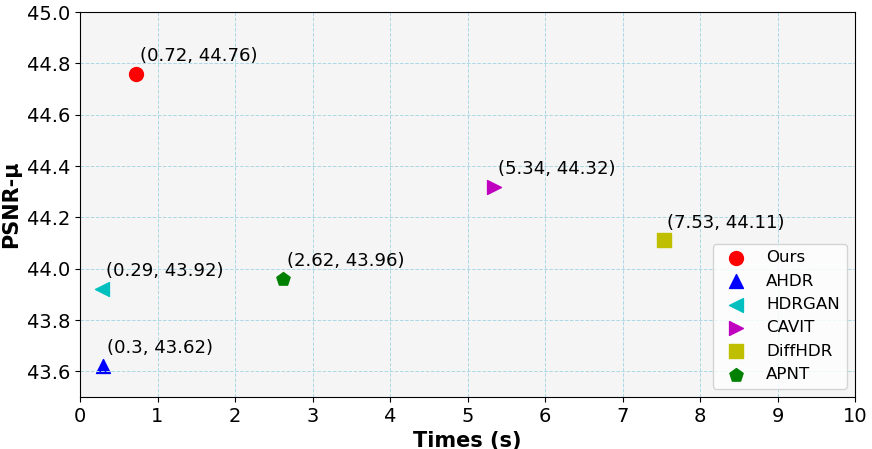}
\caption{
Average PSNR \vs inference time on the Kalantari's dataset \cite{Kalantari2017Deep}. Our method performs favorably and is 10$\times$ faster than the previous diffusion-model-based method DiffHDR \cite{10288540}.
}
\label{intur}
\vspace{-0.4cm}
\end{figure}
Multiple exposure High Dynamic Range (HDR) imaging aims to restore missing details from exposure-varied Low Dynamic Range (LDR) images. 
However, in dynamic scenes, it often causes ghosting artifacts due to object or camera movement, limiting practical applications. Researchers are actively exploring ghost-free image reconstruction methods to enable seamless capture of dynamic scenes with high dynamic ranges.

In recent years, the rise of Deep Neural Networks (DNNs) has brought significant advancements to HDR imaging field. Numerous DNN-based methods have emerged, employing Convolutional Neural Networks (CNNs) \cite{Kalantari2017Deep,yan2019attention,10446711} or Vision Transformers (ViTs) \cite{liu2022ghost, song2022selective} for HDR image reconstruction.
Despite their advancements, DNN-based approaches face challenges when essential information (\eg, content, details) for overexposed areas is missing due to object or camera movement.
%
%
%
More recently, Diffusion Models (DMs) \cite{pmlr-v37-sohl-dickstein15,NEURIPS2020_4c5bcfec} have exhibited impressive performance in image synthesis \cite{NEURIPS2021_49ad23d1,song2020score} and image reconstruction tasks (including HDR imaging) \cite{saharia2022image, 10288540,10446217}.
%
DMs achieve this by iteratively denoising Gaussian noise to generate high-fidelity images.
Yan \etal~\cite{10288540} reconstruct complete HDR images from pure Gaussian noise, yielding impressive results.
Note that, compared to previous Deep generative models \cite{niu2021hdr,9826814}, 
DMs generate a more accurate target distribution without facing issues like optimization instability or mode collapse, making them a promising choice for enhancing HDR image quality.
%
Nonetheless, 
DMs demand a substantial number of iteration steps on 
denoising models to capture intricate data details, which is time-consuming even with a high-end GPU card. For example, as shown in Fig. \ref{intur}, DiffHDR \cite{10288540} takes about 7.5 seconds to generate an HDR image on a single A100 GPU.
In this paper, our goal is to devise a DM-based method that effectively harnesses the robust distribution mapping capabilities of DMs for HDR image reconstruction.
We note that it is unnecessary to reconstruct a complete HDR image from pure noise,
because the LDR reference images already provide most of the necessary content information for HDR images.
As low-frequency information is closely related to human visual perception \cite{wang2020high}, we propose to utilize DMs to create compact low-frequency priors in the latent space. 
These priors are then integrated into the regression-based model to predict the low-frequency content of the reconstructed HDR images.

To achieve our goal, we propose the Low-Frequency Aware Diffusion Model (LF-Diff), comprising a Low-Frequency Prior Extraction Network (LPENet), a denoising network, and a Dynamic HDR Reconstruction Network (DHRNet). 
%
%
%
%
Following established methods \cite{rombach2022high,chen2024hierarchical,xia2023diffir}, we employ a two-stage training strategy for pre-train LF-Diff and DM training.
In the first stage, LPENet learns to extract a compact Low-frequency Prior Representation (LPR) from ground-truth images, guiding the DHRNet.
DHRNet consists of two modules: a Prior Integration Module (PIM) that fuses the LPR with intermediate features of DHRNet, and a Feature Refinement Module (FRM) that further processes the fused features to HDR image.
Notably, LPENet and DHRNet are optimized together in this phase.
In the second stage, we train DM to learn the compact LPR directly from LDR images. 
Since the LPR is lightweight and only used to low-frequency content for HDR imaging, our DM can estimate the LPR with extremely low computational cost, ensuring stable visual results after multiple iterations. 

The main contributions can be summarized as follows:
\begin{itemize}[leftmargin=*]
    \item We introduce LF-Diff, a straightforward and efficient DM-based method for HDR imaging.  LF-Diff leverages the capabilities of diffusion models to generate informative low-frequency priors, which are then integrated into the imaging process to enhance results.
    
    \item  We propose PIM and FRM in the DHRNet to fully exploit the LPR. While PIM efficiently fuses the LPR with intermediate features of DHRNet, FRM further processes the fused features to reconstruct high-quality HDR images.
    
    \item Extensive experiments demonstrate that the proposed LF-Diff method achieves SOTA performance in HDR imaging tasks, and produces visually appealing results that align with human visual perception. But consuming significantly fewer computational resources compared to other DM-based methods.
\end{itemize}
    
    

\section{Related Work}
\noindent\textbf{Ghost-free HDR Reconstruction.}
Traditional methods often employ motion rejection \cite{Heo2011ACCV,yan2017high}, motion registration \cite{Ward2012,Tomaszewska07}, and patch matching \cite{sen2012robust,Hu2013deghosting} approaches to align LDR images and reconstruct high-quality results. However, these methods heavily rely on the performance of preprocessing techniques and often face challenges when dealing with motions across large spatial extents.
Subsequently, DNN-based methods have emerged as the mainstream approach, thanks to their superior nonlinear expression capability. Researchers have explored various applications of DNNs in HDR imaging, devising sophisticated network architectures and model optimization schemes, such as attention \cite{chen2022attention,yan2019attention}, transformer \cite{liu2022ghost,song2022selective,Tel_2023_ICCV}, optical flow \cite{Kalantari2017Deep,peng2018deep}, GAN \cite{goodfellow2020generative,niu2021hdr,9826814}, and others \cite{yan2023unified,yan2020deep,liu2021adnet,10446711,fang2024glgnet}.
Nevertheless, when LDR images lack sufficient information due to motion or saturation, they often manifest artifacts not aligned with human visual perception, commonly referred to as ghosting.

\noindent\textbf{Diffusion Models.}
Diffusion models have recently demonstrated promising results in various low-level vision tasks, including super-resolution \cite{li2022srdiff,saharia2022image}, HDR reconstruction \cite{10288540,10446217}, colorization \cite{wang2022zero,kawar2022denoising}, and deblurring \cite{whang2022deblurring}. DiffHDR \cite{10288540} applied the conventional diffusion model paradigm to reconstruct HDR images, yielding remarkable outcomes. However, due to the high computational cost of diffusion models along with the requirement for numerous iterative steps, inference speed is limited. Despite efforts to mitigate this constraint \cite{chen2020wavegrad,DBLP:conf/iclr/SongME21,lu2022dpm}, the overall complexity remains high, especially for high-resolution images commonly encountered in HDR scenarios.
Recently, several methods \cite{rombach2022high,chen2024hierarchical,xia2023diffir} have addressed this issue by conducting diffusion modeling in the latent space. Rombach \etal \cite{rombach2022high} employed an autoencoder model to compress images into features equivalent to the image space. HI-Diff \cite{chen2024hierarchical} leverages diffusion-based vector features for assisting image deblurring. However, given the unique characteristics of HDR imaging, further exploration is needed to select an appropriate latent space for HDR reconstruction tasks.

%% file: sec/2_methods.tex
\begin{figure*}[t]
\centering
\includegraphics[width=1\textwidth,height=0.42\textwidth]{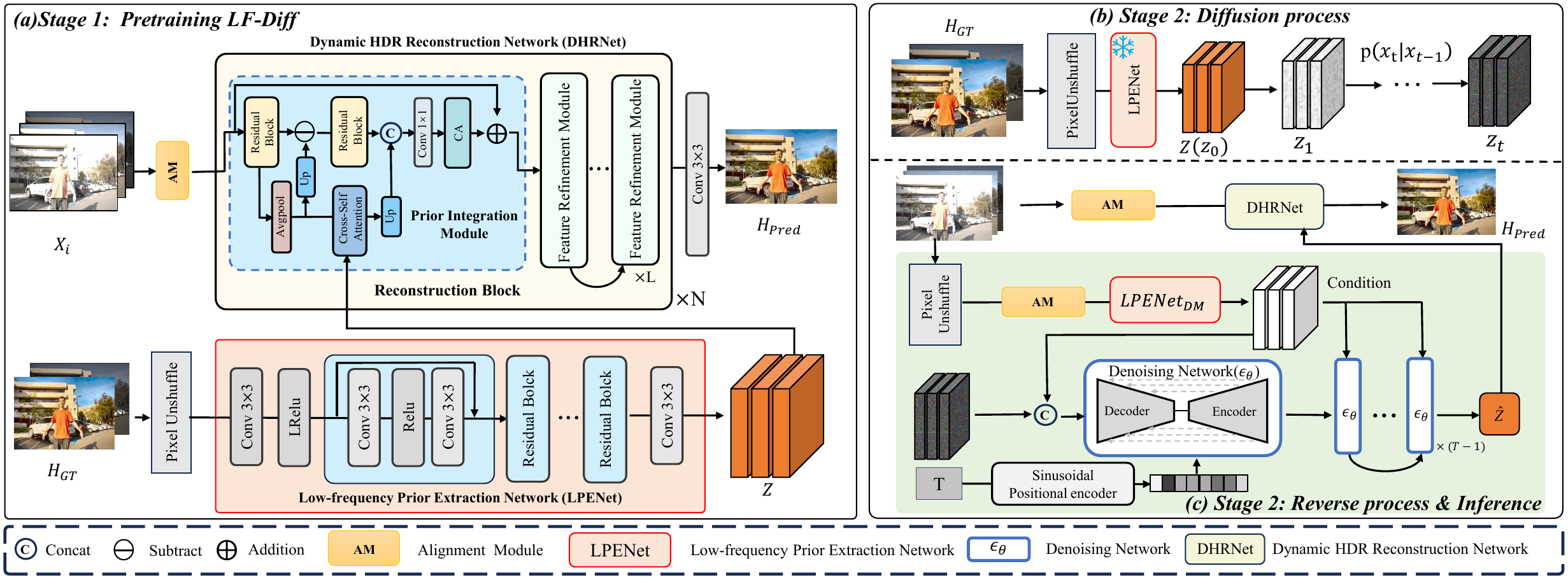}
\caption{The proposed LF-Diff comprises DHRNet, LPRNet, and a denoising network. LF-Diff undergoes two training stages: Pretrain LF-Diff (Sec. \ref{Stage One}) and DM training (Sec. \ref{Stage Two}). Notably, during the inference stage, we do not input the ground-truth image into LPENet$_{DM}$ and the denoising networks. Instead, we solely utilize the reverse process of DMs.}
\label{frame}
\end{figure*}

\section{Preliminaries}
\label{dpm}
The diffusion model introduced by Sohl-Dickstein \etal \cite{pmlr-v37-sohl-dickstein15} is inspired by nonequilibrium thermodynamics. Here, we provide a brief overview of the "variance-preserving" diffusion model from \cite{NEURIPS2020_4c5bcfec}, which encompasses diffusion processes and reverse processes.

\noindent\textbf{Diffusion Process.} 
Given a clean image distribution, the diffusion process gradually injects isotropic Gaussian noise to generate $x_t$ according to a variance schedule $\beta_1,\cdots,\beta_T\in(0,1)$. Let $\alpha_t=1-\beta_t$, $\bar{\alpha}_t=\prod^T_{i=1}\alpha_i$:
\begin{equation}
\begin{aligned}
q(x_t|x_{0}) = \mathcal{N}(x_t;\sqrt{\bar{\alpha}_t}x_{0},(1-\bar{\alpha}_t)\mathbf{I}).
\end{aligned}
\label{diffusion_process}
\end{equation}
The closed-form for this process can be represented as:
\begin{equation}
  x_t(x_0,\epsilon_t)=\sqrt{\bar{\alpha}_t}x_0+\sqrt{1-\bar{\alpha}_t}\epsilon_t,\epsilon\in(0,\mathbf{I}),
\label{x0xt}
\end{equation}
where $x_t$ represents the output at time $t$, $x_0$ is the initial clean image, and $\epsilon_t$ denotes isotropic Gaussian noise.

\noindent\textbf{Reverse Process.} 
The reverse process attempts to approximate the data distribution $q(x_0)$ by a Markovian process, starting with random Gaussian noise $x_T=\mathcal{N}(x_T;\mathbf{0},\mathbf{I})$:
\begin{equation}
\begin{split}
q(x_{t-1}|x_{t},x_0) = \mathcal{N}(x_{t-1};\Tilde{\mu}_t(x_t,x_0),{\Tilde{\beta}_t}\mathbf{I}),
\end{split}
\label{reverse_process}
\end{equation}
where $\Tilde{\mu}_t$ and $\Tilde{\beta}_t$ are distribution parameters. 
Since the true reverse process Eq.\eqref{reverse_process} relies on $q(x_0)$ and is intractable, the neural network $f_{\theta}$ serves as the denoiser to estimate $p_{\theta}(x_{t-1}|x_t, t)$ instead of $q(x_{t-1}|x_t, x_0)$:
\begin{align}
\begin{split}
p_\theta(x_{t-1}|x_t,t) &= \mathcal{N}(x_{t-1};\mu_{\theta}(x_t,t),\Sigma_{\theta}(x_t,t)) \\
&= q(x_{t-1}|f_{\theta}(x_t,t)),
\end{split}
\label{reverse_process1}
\end{align}
where $\mu_{\theta}$ and $\Sigma_{\theta}$ are the parameters to be estimated in the reverse process.
Let fixed variances ($\Sigma_{\theta}(x_t,t)=\Tilde{\beta}_t\mathbf{I}$), the optimization objective can be defined as:
\begin{align}
\label{Lnoise}
    \mathbb{E}_{t,x_0,\epsilon_t}[\Vert\epsilon_t-f_{\theta}(\sqrt{{\bar{\alpha}_t}}x_0+\sqrt{1-\bar{\alpha}_t}\epsilon_t,t)\Vert^2].
\end{align}

\section{Methodology}
As depicted in Fig. \ref{frame}, we introduce the proposed LF-Diff framework, which consists of two training stages.
During the first stage, we pre-train LF-Diff to extract accurate low-frequency prior representations from ground truths using LPENet, while DHRNet learns how to utilize the above priors to reconstruct HDR images. During the second stage, we train the DM to accurately predict the low-frequency prior representations directly from LDR images, and jointly optimize it with DHRNet to generate final high-quality HDR images.

\subsection{Stage One: Pretrain LF-Diff}
\label{Stage One}
In the first stage, our objective is to train the LPENet how to extract accurate low-frequency prior features from ground truth and embed this representation into the DHRNet to guide the reconstruction process.
As illustrated in Fig. \ref{frame} (a), we utilize the Prior Integration Module (PIM) to  incorporate these prior features into the DHRNet via the cross-attention mechanism. Within DHRNet, multiple Feature Refinement Modules (FRM) are stacked to directly learn the mapping relationship from input features to the target image distribution, while PIM leverages prior features to assist in recovering the low-frequency content of the reconstructed images. Next, we elaborate on above descriptions.

\textbf{Low-frequency Prior Extraction Network.}
The LPENet has a straightforward structure, as shown in Fig. \ref{frame} (a), comprising several residual blocks designed to extract the Low-frequency Prior Representation (LPR).
Starting with the ground truth image $I_{gt} \in \mathbb{R}^{H \times W \times 3}$, we initially perform a tonemapping operation $\mathcal{T}(\cdot)$ to obtain the LDR domain ground truth image $\mathcal{T}(I_{gt})$. We concatenate these images along the channel dimension and employ the PixelUnshuffle operation to downsample, generating the input for LPENet. Subsequently, LPENet extracts the LPR $z \in \mathbb{R}^{\frac{H}{k} \times \frac{W}{k} \times 3}$ as follows:
\begin{equation}
\label{LPEN}
\small
z=\mathrm{LPENet}\left(\operatorname{PixelUnshuffle}\left(\operatorname{Concat}\left(I_{gt}, \mathcal{T}(I_{gt})\right)\right)\right),
\end{equation}
where $\mathcal{T}(x) = \frac{log(1+\mu x)}{log(1+\mu)}$ denotes the tonemapping operator, $\mu =5000$ is a predefined constant parameter that controls the degree of compression applied to the input signal, $k$ denotes the sampling multiple of PixelUnshuffle.

\textbf{Dynamic HDR Reconstruction Network.}
As illustrated in Fig. \ref{frame} (a), DHRNet consists of multiple stacked Reconstruction Blocks, with each block comprising one Prior Integration Module (PIM) and $N$ Feature Refinement Module (FRM).
PIM fuses the LPR with intermediate features of DHRNet, we will give a detailed description in the following part.
FRM is designed based on the understanding that natural images encompass diverse frequencies, since high and low frequencies serve distinct roles in image encoding.

\begin{figure}[t]
\centering
\includegraphics[width=1\linewidth]{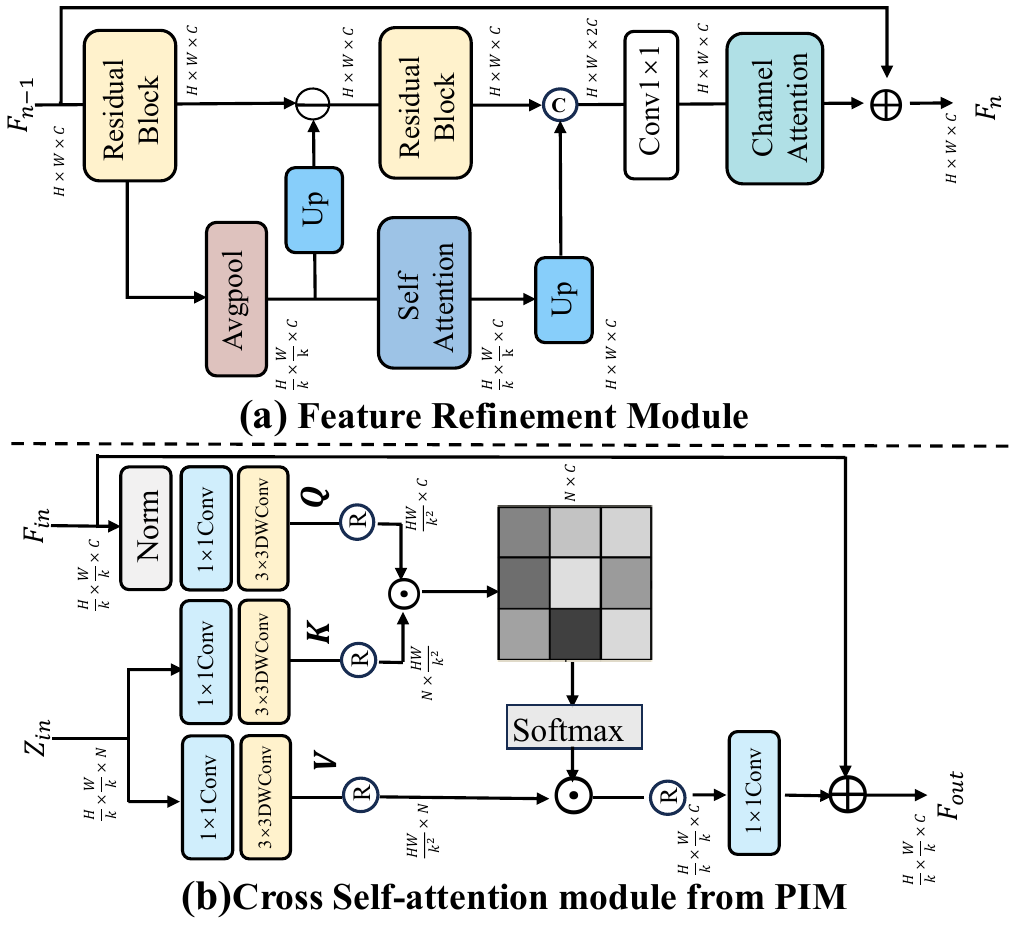}
\caption{The architecture of (a) Feature Refinement Module (FRM) 
and (b) Cross-attention mechanism.
}
\vspace{-0.4cm}
\label{hpb}
\end{figure}

In Fig. \ref{hpb} (a), the structure of FRM involves several blocks. Specifically, we begin by processing the features through a residual block to obtain $F_{n-1}$. Subsequently, we decouple these features into two components: high-frequency features $F_{n-1}^{\text {high}}$, which remain resolution-invariant, and low-frequency features $F_{n-1}^{\text {low}}$ with reduced resolution, following the approach described in \cite{lu2022transformer}:
\begin{equation}
\begin{aligned}
& F_{n-1}^{\text {low }}=\operatorname{Avgpool}\left(F_{n-1}, k\right), \\
& F_{n-1}^{\text {high }}=F_{n-1}-\text { Upsample }\left(F_{n-1}^{\text {low }}\right),
\end{aligned}
\end{equation}
where $k$ denotes the kernel size of the avgpooling layer, $Upsample$ denotes the upsampling operator.
In essence, low-frequency features capture global dependencies within the input (image/features) and do not necessitate a high-resolution feature map, but require global attention. On the other hand, high-frequency features capture detailed local information, requiring a high-resolution feature map, and only can be achieved through local operators. Therefore, we employ the residual block (RB) to process $F_{n-1}^{\text {high}}$ and utilize the self-attention (SA) from \cite{zamir2022restormer} to handle $F_{n-1}^{\text {low}}$.
Subsequently, we fuse the low- and high-frequency features to preserve the initial details, resulting in the feature $F_{n-1}^{\prime}$. This operation can be represented as follows:
\begin{equation}
\small
F_{n-1}^{\prime}=Concat\left[R B\left(F_{n-1}^{\text {high}}\right), Upsample\left(S A\left(F_{n-1}^{\text {low}}\right)\right)\right].
\end{equation}
Considering that $F_{n-1}^{\prime}$ is a concatenation of two features, a $1\times 1 $ $Conv$ is employed to effectuate a reduction in the number of channels. This downsizing step is followed by a channel attention module, designed to accentuate those channels exhibiting heightened activation values.

In order to effectively merge the prior feature derived from the LPENet with the intermediate feature produced by DHRNet, we replaced the self-attention mechanism in the low-frequency branch of FRM with a carefully designed cross-attention mechanism, resulting in the Prior Integration Module (PIM).
As depicted in Fig. \ref{frame} (a), a PIM is positioned before each reconstruction block. Within each PIM, cross-attention computations are performed between the prior and intermediate features to facilitate feature fusion. This module enables the aggregation of information from the prior feature into the features of DHRNet.

Specifically, as illustrated in Fig. \ref{hpb} (b), our cross-attention module receives two types of input: the intermediate feature $\mathbf{F} \in \mathbb{R}^{\hat{H} \times \hat{W} \times \hat{C}}$ and the prior feature $z \in \mathbb{R}^{\hat{H} \times \hat{W} \times \hat{N}}$.
We project $\mathbf{F}$ into a query vector $\mathbf{Q}=W_d^Q W_c^Q \mathbf{F}$ using point-wise $1\times 1$ convolution and depth-wise $3\times 3$ convolution with weights $W_d^Q$ and $W_c^Q$. Similarly, $z$ is transformed into key $\mathbf{K}$ and value $\mathbf{V}$ through analogous operations.
Next, we reshape these projections into matrices suitable for attention computation: $\hat{\mathbf{Q}} \in \mathbb{R}^{\hat{H} \hat{W} \times \hat{C}}$, $\hat{\mathbf{K}} \in \mathbb{R}^{  \hat{N} \times \hat{H} \hat{W}}$, and $\hat{\mathbf{V}} \in \mathbb{R}^{\hat{H} \hat{W} \times \hat{N}}$. We then calculate a more computationally efficient attention map $A \in \mathbb{R}^{\hat{N} \times \hat{C}}$ by performing dot product between $\hat{\mathbf{Q}}$ and $\hat{\mathbf{K}}$.
The process can be described as follows:
\begin{equation}
\hat{\mathbf{F}}=W_c \hat{\mathbf{V}} \cdot \operatorname{Softmax}(\hat{\mathbf{K}} \cdot \hat{\mathbf{Q}} / \gamma)+\mathbf{F},
\end{equation}
where $\gamma$ is a trainable parameter. Notably, we do not implement multi-head attention in PIM.


\textit{\textbf{Training Strategy.}}
As proposed in \cite{Kalantari2017Deep}, we initially transform the provided set of input LDR images $\left\{L_1, L_2, \ldots, L_N\right\}$ into their corresponding HDR versions $H_i$ using gamma correction. Subsequently, we concatenate each $L_i$ with its corresponding $H_i$ along the channel axis to generate six-channel input tensors $X_i = [L_i, H_i]$.
Similar to \cite{yan2019attention,liu2022ghost,10288540}, we employ an alignment module \cite{yan2019attention} to process the input LDR images, align the features implicitly, and feed them into DHRNet. 
Then, the LPR feature $z$  extracted from LPENet is injected into DHRNet via PIM. The reconstructed HDR image $\hat{H}$ is generated by:
\begin{equation}
\label{output}
\hat{H}=\operatorname{Conv}_{3 \times 3}\left(\operatorname{DHRNet}\left(X_i, z\right)\right).
\end{equation}
Following previous approaches \cite{yan2019attention,liu2022ghost}, we utilize both tonemapped per-pixel loss and perceptual loss as the image reconstruction loss function $\mathcal{L}_r$. This dual-loss strategy optimizes both pixel-level accuracy and high-level feature representations in the generated HDR results:
\begin{equation}
\scriptstyle
\label{Lr}
\left.\mathcal{L}_r=\| \mathcal{T}(H)-\mathcal{T}(\hat{H})\right)\left\|_1+\lambda\right\| \phi_{i, j}(\mathcal{T}(H))-\phi_{i, j}(\mathcal{T}(\hat{H})) \|_1,
\end{equation}
where $\phi_{i,j}(\cdot)$ signifies the j-th convolutional feature extracted from the VGG19 network after the i-th maxpooling operation, and $\lambda=$1e-2 is a hyperparameter that balances the contribution of the each component.

\subsection{Stage Two: Diffusion Models for HDR Imaging}
\label{Stage Two}
After the above-mentioned learning procedure, we now possess LDR images and their corresponding LPR $z$. In the second stage (Fig. \ref{frame} (b, c)), our objective is to efficiently harness the powerful distribution estimation capability of the DMs. Specifically, we utilize the LPENet from stage one to capture the LPR as the denoising target of the DM. The DM will learn how to extract accurate LPR from the LDR inputs and perform joint optimization with DHRNet.


\textbf{Diffusion Model for LRP Learning.}
After capturing the LPR $z \in \mathbb{R}^{\frac{H}{k} \times \frac{W}{k} \times N}$ from the pretrained LPENet, we transform the clean LPR feature $z$ into a noisy version $z_T$ using Eq.\eqref{x0xt}, which can be described as:
\begin{equation}
q\left(z_T \mid z\right)=\mathcal{N}\left(z_T ; \sqrt{\bar{\alpha}_T} z,\left(1-\bar{\alpha}_T\right) \mathbf{I}\right),
\end{equation}
where $T$ represents the total number of iterations, while $\bar{\alpha}_t$ and $\alpha$ denote the pre-defined variance schedule.

In the reverse process, we iteratively generate the LPR from a pure Gaussian distribution, starting from $z_T$ and moving backward to $z_0$, which utilizes the posterior distribution described in Eq. \eqref{reverse_process}.
Following previous work \cite{10288540,saharia2022image}, we utilize a neural network to estimate $p_{\theta}(x_{t-1}|x_t, t)$ instead of $q(x_{t-1}|x_t, x_0)$ for each step.
Specifically, we first use $LPENet_{DM}$ to obtain a conditional feature $D \in \mathbb{R}^{\frac{H}{k} \times \frac{W}{k} \times N}$ from aligned LDR feature:
\begin{equation}
\label{lpendm}
\mathbf{D}=\operatorname{LPENet}_{DM}\left(AM(\operatorname{PixelUnshuffle}\left(X_i\right)\right)),
\end{equation}
where $\operatorname{LPENet}_{DM}$ maintains a similar architecture to $\operatorname{LPENet}$ but with a modified input dimensionality for its first convolution layer, $AM(\cdot)$ denotes alignment module from \cite{yan2019attention}.
The denoising neural network predicts the noise based on both $z_t$ and the derived conditional feature $D$, \ie, $\boldsymbol{\epsilon}_\theta\left(Concat(z_t, \mathbf{D}), t\right)$. 
Upon substituting this estimated noise term into the equation governing the reverse process Eq. \eqref{reverse_process1}, we arrive at the following sampling formula:
\begin{equation}
\label{sample}
\boldsymbol{z}_{t-1}=\frac{1}{\sqrt{\alpha_t}}\left(\boldsymbol{z}_t-\frac{1-\alpha_t}{\sqrt{1-\bar{\alpha}_t}} \boldsymbol{\epsilon}_\theta\left(z_t, \mathbf{D}, t\right)\right)+\sqrt{1-\alpha_t} \boldsymbol{\epsilon}_t,
\end{equation}
where $\boldsymbol{\epsilon}_t \sim \mathcal{N}(0, \mathbf{I})$. By iteratively sampling $z_t$ using Eq. \eqref{sample} $T$ times, we progressively reconstruct the predicted LPR representation $z_0$ (\ie $\hat{z}$)  $\in \mathbb{R}^{\frac{H}{k} \times \frac{W}{k} \times N}$.

\textbf{Training Strategy.}
\label{stage2train}
Traditional DMs solely learn the probability distribution by optimizing the weighted variational bound (Eq. \eqref{Lnoise}), resulting in a slight deviation between the predicted prior and the actual prior. Integrating the DM directly with DHRNet might lead to a misalignment issue, thereby hampering the overall image processing performance. To address this, we joint training the DM and DHRNet.
During each training iteration, we first sample the noise sample $z_t$ through the diffusion process (Eq. \eqref{x0xt}). Given that our denoising neural network is lightweight, we use a reverse process for $S$ iterations based on the DDIM strategy \cite{DBLP:conf/iclr/SongME21} to generate the predicted prior feature $\hat{z}$. This $\hat{z}$ guides DHRNet via PIM. The loss function in the second stage consists of the reconstruction loss $\mathcal{L}_r$ (Eq. \eqref{Lr}) and the diffusion loss $\mathcal{L}_{diff}$:
\begin{equation}
\begin{aligned}
\mathcal{L}_{\text {diff }} & =E_{z_0, t, \epsilon_t \sim \mathcal{N}(\mathbf{0}, \mathbf{I})}\left[\left\|\epsilon_t-\epsilon_\theta\left(z_t, D, t\right)\right\|^2\right] \\
& +\left\|\hat{z}-z\right\|_1.
\end{aligned}
\label{Ldiff}
\end{equation}

\subsection{Inference}
In the inference stage, $LPENet_{DM}$ extracts a conditional feature $\mathbf{D}$ from aligned LDR feature (Eq. \eqref{lpendm}), and we randomly sample a Gaussian noise $\hat{z}_T$. Then, denoising neural network utilizes the $\hat{z}_T$ and $\mathbf{D}$ to estimate LPR after $S$ iterations (Eq. \eqref{sample}) based on DDIM \cite{DBLP:conf/iclr/SongME21}. After that, DHRNet exploits the LPR to restore HDR images as Eq. \eqref{output}. More details about the DM training and the model inference can be found in the supplementary material.

%% file: sec/3_experiments.tex
\begin{figure*}[tb]
\centering
\includegraphics[width=1\linewidth,height=0.34\textwidth]{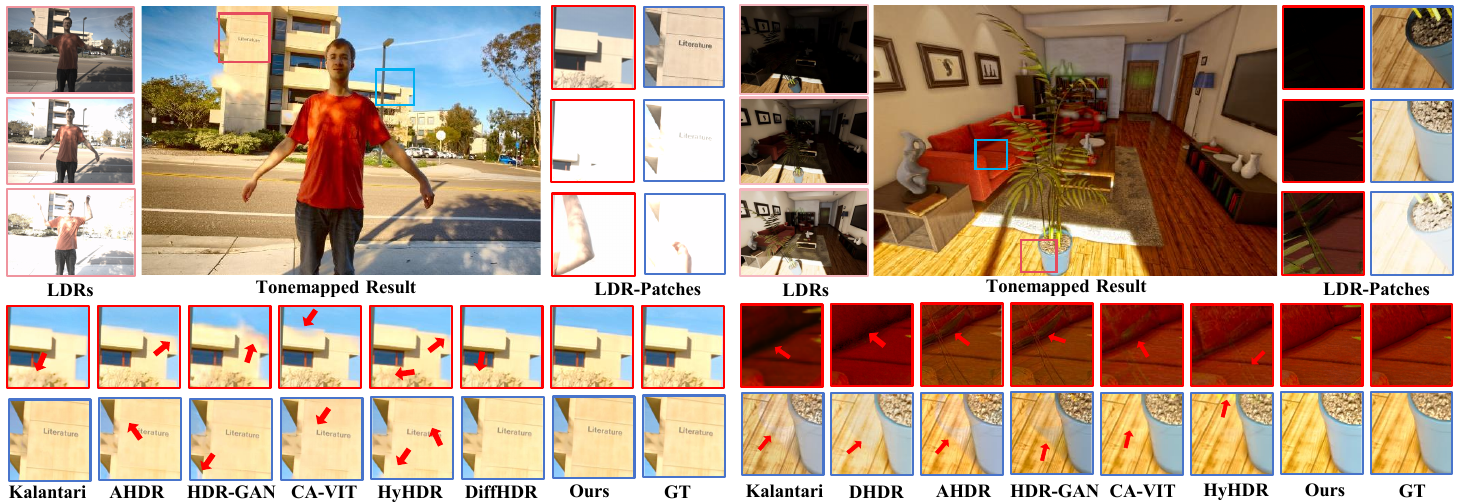}
\footnotesize (a) Example from Kalantari \etal's dataset \cite{Kalantari2017Deep} \qquad\qquad\qquad\qquad\qquad\qquad\qquad\qquad (b) Example from Hu \etal's dataset \cite{hu2020sensor}
\caption{Visual comparisons are conducted on testing data from various datasets, focusing on zoomed-in local areas of the HDR images estimated by our method and the compared techniques. Our model demonstrates the ability to generate HDR images of superior quality.
}
\label{comparewgt}
\vspace{0cm}
\end{figure*}
\section{Experiments}
\subsection{Experimental Settings}
\noindent\textbf{Datasets.} 
All methods are trained using two publicly available datasets, employing identical training settings: Kalantari's dataset \cite{Kalantari2017Deep} and Hu's dataset \cite{hu2020sensor}. Kalantari's dataset consists of 74 samples for training and 15 for testing, all captured under authentic environmental conditions. Each sample comprises three LDR images with exposure variations of \{-2, 0, +2\} or \{-3, 0, +3\}. In contrast, Hu's dataset is a synthetic dataset designed to emulate sensor realism, generated through a game engine. This dataset contains images captured at three distinct exposure levels \{-2, 0, +2\}, with our primary focus on dynamic scene images within this collection. Following the settings outlined in \cite{hu2020sensor}, we allocate the initial 85 samples for training, reserving the remaining 15 for testing. Moreover, to further validate the model's generalizability, we incorporate Sen \etal.’s dataset \cite{sen2012robust} and Tursun \etal’s dataset \cite{Tursun2016data} exclusively for qualitative assessment, as these datasets lack ground truths.


\noindent\textbf{Evaluation Metrics.}
To conduct a quantitative comparison, we utilize five objective metrics: PSNR-$\mu$, SSIM-$\mu$, PSNR-L, SSIM-L, and HDR-VDP-2 \cite{Mantiuk2011HDR}. Here, the subscripts $\mu$ and L denote that these metrics are computed in the tonemapped and linear domains, respectively.

\begin{figure*}[tb]
\centering
\includegraphics[width=1\linewidth,height=0.34\textwidth]{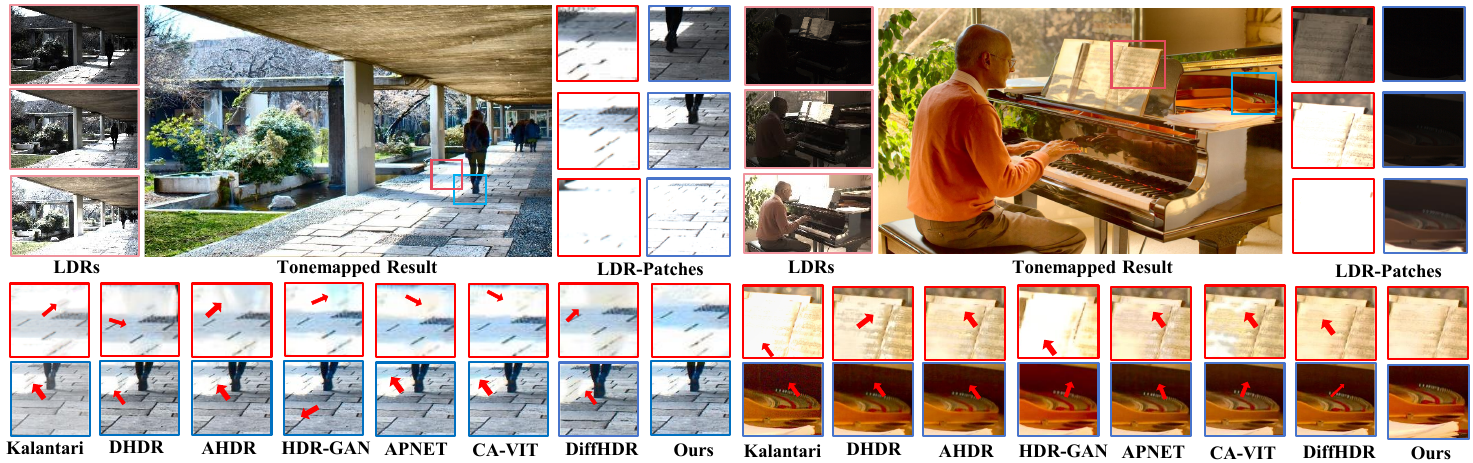}
\footnotesize (a) Example from Tursen \etal's dataset \cite{Tursun2016data} \qquad\qquad\qquad\qquad\qquad\qquad\qquad\qquad (b) Example from Sen \etal's dataset \cite{sen2012robust}
\caption{Visual comparisons on the datasets without ground truth.}
\label{comparewogt}
\end{figure*}
\noindent\textbf{Implementation Details.}
Our implementation is conducted using PyTorch, and each training stage converges after 300 epochs on four NVIDIA A100 GPUs. We employ the Adam optimizer with an initial learning rate of 1e-4, which decays by a factor of 0.1 after every 50 epochs. The training dataset is processed by cropping $128\times 128$ patches with a stride of 64, and the batch size is set to 64.
For LF-Diff, we adopt a variant of the commonly used U-Net architecture from \cite{luo2023refusion} as the denoising network, with 3 levels of blocks \{2, 2, 2\}. During training, the time step $T$ is set to 200, and the implicit sampling step $S$ is set to 10 for both training and inference phases to achieve efficient restoration.
In DHRNet, the parameters $N_i \in \{L_1,L_2,L_3\}$ are set to \{3, 3, 3\}, and the channel $C$ is set to 60. The number of attention heads for the self-attention branch at the same level is set to \{6, 6, 6\}, and the channel expansion factor in FFN is 2.66. The kernels for avgpool in PIM and FRM are set to 4 and 2, respectively.
For LPENet, it comprises 4 residual blocks, with a Pixelunshuffle downsampling factor of 4 and the output channel of 3. 

\subsection{Comparison with the State-of-the-art Methods}
In this section, we evaluate the performance of the proposed LF-Diff method and present experimental results to validate its reconstruction performance compared to state-of-the-art techniques. Specifically, we evaluate LF-Diff against two patch-based methods \cite{sen2012robust,Hu2013deghosting}, a optical flow-based method \cite{Kalantari2017Deep}, five CNN-based methods \cite{Wu_2018_ECCV,yan2019attention,niu2021hdr,liu2021adnet,chen2022attention}, 
two ViT-based methods \cite{liu2022ghost,yan2023unified}, along with a diffusion-based method \cite{10288540}. It's worth noting that all deep learning methods employ the same training dataset and settings for consistency.

\begin{table*}[htbp]\small
    \centering
    \setlength{\tabcolsep}{2pt}
    
    \scalebox{0.80}{
    \begin{tabular}{c|c|cccccccccccc}
        \toprule
        {Datasets} & {Models} & {Sen}\cite{sen2012robust}  & {Hu}\cite{Hu2013deghosting} & {Kalantari}\cite{Kalantari2017Deep} & {DHDR}\cite{Wu_2018_ECCV} & {AHDR}\cite{yan2019attention} & {NHDRR}\cite{yan2020deep}& {HDRGAN}\cite{niu2021hdr} & {APNT}\cite{chen2022attention}  & {CA-ViT}\cite{liu2022ghost} & {HyHDR}\cite{yan2023unified} & {DiffH}\cite{10288540} & { Ours}\\
        \midrule
        \multirow{5}{*}{Kalantari} & PSNR-$\mu$ & 40.95 & 32.19 & 42.74 & 41.64 & 43.62 &42.41& 43.92 & 43.94  & 44.32 & 44.64 & 44.11 & \textbf{44.76}\\
        & PSNR-L & 38.31 & 30.84 & 41.22 & 40.91 & 41.03 & 41.08 & 41.57 & 41.61  & 42.18 & 42.47 & 41.73  & \textbf{42.59}\\
        & SSIM-$\mu$ & 0.9805 & 0.9716 & 0.9877 & 0.9869 & 0.9900 & 0.9887 & 0.9905 & 0.9898  & 0.9916 & 0.9915 & 0.9911 &\textbf{0.9919}\\
        & SSIM-L & 0.9726 & 0.9506 & 0.9848 & 0.9858 & 0.9862 & 0.9861 & 0.9865 & 0.9879  & 0.9884 & 0.9894 & 0.9885 & \textbf{0.9906}\\
        & HDR-VDP-2 & 55.72 & 55.25 & 60.51 & 60.50 & 62.30 & 61.21 & 65.45 & 64.05 & 66.03 & 66.05 & 65.52 & \textbf{66.54}\\
        \bottomrule
    
        {Datasets} & {Models} & {Sen}\cite{sen2012robust}  & {Hu}\cite{Hu2013deghosting} & {Kalantari}\cite{Kalantari2017Deep} & {DHDR}\cite{Wu_2018_ECCV} & {AHDR}\cite{yan2019attention} & {NHDRR}\cite{yan2020deep}& {HDRGAN}\cite{niu2021hdr} & {APNT}\cite{chen2022attention} & {ADNet}\cite{liu2021adnet} & {CA-ViT}\cite{liu2022ghost} & {HyHDR}\cite{yan2023unified}  & { Ours}\\
        \midrule
        \multirow{5}{*}{Hu} & PSNR-$\mu$ & 31.48 & 36.56 & 41.60 & 41.13 & 45.76 & 45.15 & 45.86 & 46.41 & 46.79 & 48.10 & {48.46}  & \textbf{48.74}\\
        & PSNR-L & 33.58 & 36.94 & 43.76 & 41.20 & 49.22 & 48.75 & 49.14 & 47.97 & 50.38 & 51.17 & {51.91} & \textbf{52.10} \\
        & SSIM-$\mu$ & 0.9531 & 0.9824 & 0.9914 & 0.9870 & 0.9956 & 0.9956 & 0.9945 & 0.9953 & 0.9908 & 0.9947 & {0.9959} & \textbf{0.9968} \\
        & SSIM-L & 0.9634 & 0.9877 & 0.9938 & 0.9941 & 0.9980 & 0.9981 & 0.9989 & 0.9986 & 0.9987 & 0.9989 & {0.9991} & \textbf{0.9993}  \\
        & HDR-VDP-2 & 66.39 & 67.58 & 72.94 & 70.82 & 75.04 & 74.86 & 75.19 & 73.06 & 76.21 & 77.12 & {77.24} & \textbf{77.35} \\
        \bottomrule
    \end{tabular}}
    \caption{The evaluation results on Kalantari’s \cite{Kalantari2017Deep} and Hu’s \cite{hu2020sensor} datasets. The best results are highlighted in \textbf{Bold}.}
    \vspace{-0.1cm}
    \label{quantitative results}
\end{table*}
\begin{table*}[t]\small
\begin{center}
\renewcommand{\arraystretch}{1.3}
 
  \scalebox{0.8}{
  \begin{tabular}{c|cccccccccc}
  \hline
  Methods &Sen \cite{sen2012robust} & Hu \cite{Hu2013deghosting}& Kalantari \cite{Kalantari2017Deep} &  AHDR \cite{yan2019attention}& NHDRR \cite{yan2020deep}& HDRGAN \cite{niu2021hdr}& APNT \cite{chen2022attention}&  CA-ViT \cite{liu2022ghost} &DiffHDR \cite{10288540}& Ours\\
  Environment &(CPU)&(CPU) &(CPU+GPU)&(GPU)&(GPU)&(GPU)&(GPU)&(GPU)&(GPU)&(GPU) \\
     \hline
     Times(s) & 61.81s & 79.77s & 29.14s & 0.30s & 0.31s& 0.29s& 2.62s& 5.34s& 7.53s& 0.72s \\
     Para.(M) & - & - & 0.30M & 1.24M & 38.10M&2.56M& 37.98M& 1.22M& 74.99M& 7.48M \\
     \hline
\end{tabular}}
\caption{Average runtime performance of various methods on the testing set with dimensions of $1000 \times 1500$.}
\label{stime}
\end{center}
\vspace{-0.5cm}
\end{table*}

\vspace{1mm}
\noindent\textbf{Datasets w/ Ground Truth.}
The quantitative outcomes for LF-Diff on two datasets are presented in Tab. \ref{quantitative results}. Our method is compared with various state-of-the-art approaches using the testing data from \cite{Kalantari2017Deep} and \cite{hu2020sensor}, which includes challenging samples characterized by saturated backgrounds and foreground motions. All quantitative results are averaged across testing images. 
Notably, LF-Diff exhibits a significant improvement over other leading methods, exceeding the performance of the DM-based method DiffHDR \cite{10288540}, by 0.65 dB and 0.86 dB in PSNR-$\mu$ and PSNR-L metrics, respectively, based on Kalantari's dataset \cite{Kalantari2017Deep}. Moreover, LF-Diff showcases superior performance compared to the runner-up method, HyHDR \cite{yan2023unified}, with improvements of 0.28 dB and 0.19 dB in PSNR-$\mu$ and PSNR-L metrics, respectively, on Hu's dataset \cite{hu2020sensor}.


As shown in Fig. \ref{comparewgt} (a) and (b), the datasets present some challenging samples due to information loss in the LDR images. Most existing approaches struggle in these areas, producing ghosting artifacts due to large motions and occlusions.
Kalantari's method \cite{Kalantari2017Deep} and DHDR \cite{Wu_2018_ECCV} struggle with background motion due to error-prone alignments (\eg, optical flow), resulting in undesirable ghosting. While HDR-GAN \cite{niu2021hdr} exhibits noticeable ghosting artifacts around the balcony area and introduces erroneous color information, AHDR \cite{yan2019attention} aligns LDR images using convolutional spatial attention. However, it unintentionally suppresses valuable contextual information and encounters difficulties with significant motions in over/underexposed scenes. CA-ViT \cite{liu2022ghost}, relying on patch-based sampling, produces noticeable blocky ghosting.
In contrast, aided by the DM, both Diff-HDR \cite{10288540} and our LF-Diff generate HDR images aligned with human perception. Notably, our approach outperforms Diff-HDR with faster inference speeds and a smaller computational overhead.

\vspace{1mm}
\noindent\textbf{Datasets w/o Ground Truth.}
To assess the generalization capability of our proposed HDR imaging method, we evaluated the performance of the model trained on Kalantari \etal's Dataset \cite{Kalantari2017Deep} by testing it on datasets from Sen \etal \cite{sen2012robust} and Tursun \etal \cite{tursun2016objective}, which lack ground truths.
In Fig. \ref{comparewogt} (a), numerous current methods encounter difficulties in restoring  recover the large saturated region and large motion. Conversely, in Fig. \ref{comparewogt} (b), while complete elimination of ghosting remains elusive for any method, our approach notably enhances image sharpness and detail compared to preceding techniques.

\vspace{1mm}
\noindent\textbf{Computational Budgets.} We also conducted comparisons regarding model parameters and inference times with previous approaches. As illustrated in Table \ref{stime}, patch match-based methods \cite{sen2012robust,Hu2013deghosting} exhibit significantly longer inference times due to their CPU-based computation. Kalantari \cite{Kalantari2017Deep} requires considerable time for initial optical flow alignment. NHDRRNet \cite{yan2020deep}, employing a U-shaped network architecture, demonstrates shorter inference times but substantially higher parameter counts compared to other methods. CA-ViT \cite{liu2022ghost} has a large number of standard transformer blocks leading to high computational cost despite fewer parameters. DiffHDR \cite{10288540} has high inference time and parameter count due to reconstructing HDR from pure noise. In comparison, our method effectively leverages the powerful distribution estimation capability of DM with orders of magnitude fewer parameters and computations.

\subsection{Ablation Study}
In this section, we investigate the impact of various designs within our proposed method. All experiments are conducted using the Kalantari's dataset \cite{Kalantari2017Deep}. 

\begin{figure}[tb]
\centering
\includegraphics[width=1.0\linewidth]{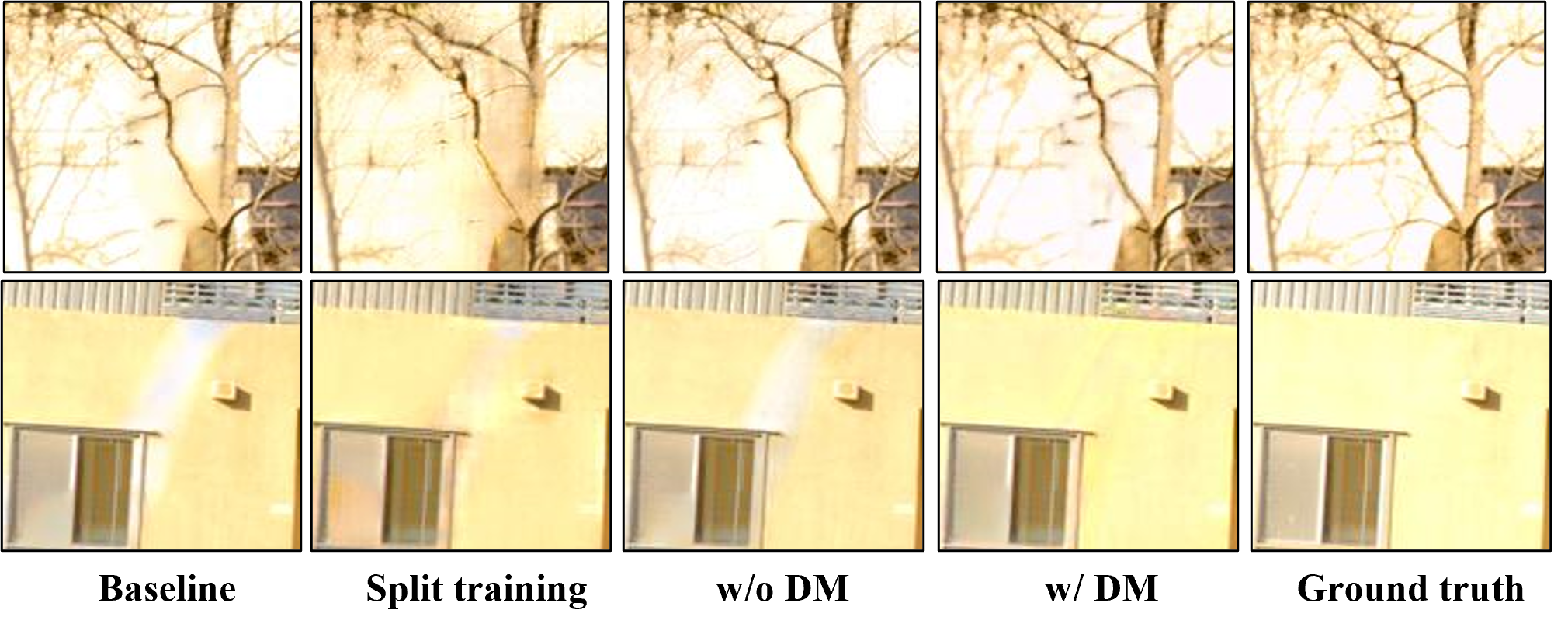}
\caption{Qualitative results of our ablation study.}
\label{ablationpng}
\end{figure}

\vspace{1mm}
\noindent\textbf{Effects of Diffusion Prior.}
As shown in Tab. \ref{abdp}, we establish a regression-based baseline model without LPR generation. In this configuration, the PIM in DHRNet is replaced with FRM, and HDR images are reconstructed through a regression-based approach. 
LF-Diff$_{s1}$ is a pre-trained model in the first stage utilizing ground-truth images to provide LPR. Compared to the baseline, this brings a 8.7dB PSNR-L improvement, demonstrating that an accurate compact prior can greatly enhance results.
Variant LF-Diff$_{s2}$ w/o DM does not use the DM, but instead directly learns the LPR using LPENet. This achieves a 0.16dB PSNR gain over the baseline. 
When learning the LPR based on the DM, LF-Diff$_{s2}$ w/ DM further improves PSNR by 0.24dB compared to variant LF-Diff$_{s2}$ w/o DM, This indicates that the DM demonstrates superior capability in accurately estimating the distribution for LPR prediction. Additionally, compared to DiffHDR, which requires over 70M Params for estimating complete images, LF-Diff only adds 2.39M Params over the baseline.
Moreover, we present visual comparisons of the baseline (without the prior feature) and LF-Diff variants in Fig. \ref{ablationpng}. It can be observed that performing DM in a compact latent space to predict LPR effectively mitigates ghosting issues.

\noindent\textbf{Effects of Joint Training Strategy.} 
We conducted an ablation study on the joint training strategy. Under this strategy, only the DM is optimized in the second stage, referred to as Split-Training. Specifically, we first utilize the pre-trained LPENet to generate the prior feature $z$ from the ground truth, and then apply the training objective defined in Eq. \eqref{Ldiff} to independently train the DM. Subsequently, the trained DM is directly integrated with the DHRNet for evaluation. It's worth noting that the DHRNet in the second stage utilizes pre-trained weights from the stage one without additional training. It is evident that LF-Diff significantly outperforms Split-Training by 1.51 dB in terms of PSNR value. This comparison underscores the effectiveness of joint training strategy over Split-Training.

\noindent\textbf{Sampling Step.}
In addition to the lightweight DM structure and compact feature dimensions, our method achieves greater computational efficiency by employing a smaller sampling step, denoted as S, in the reverse process based on DDIM \cite{DBLP:conf/iclr/SongME21}.The performance of our method is demonstrated with different values of S ranging from 5 to 100, as presented in Tab. \ref{absampletime}. Variations in sampling steps, when S is set to a value smaller than the training configuration, a noticeable performance degradation occurs. On the other hand, when S is larger than the training setting, the impact on performance is minimal, particularly with the SSIM metric remaining consistently unchanged. While a larger sampling step is known to enhance the visual quality of images in diffusion-based methods \cite{10288540,saharia2022image}, in our case, it primarily increases inference time. This observation suggests that larger sampling steps, such as S = 1000 in DiffHDR \cite{10288540} and SR3 \cite{saharia2022image}, may not be essential for diffusion-based HDR reconstruction and other related low-level vision tasks when using our proposed framework.

\begin{table}[]\small
\centering
\scalebox{0.77}{
\begin{tabular}{c|ccccccc}
\toprule
\textbf{Method}   & \textbf{GT} & \textbf{DM} & \begin{tabular}[c]{@{}c@{}}\textbf{Joint} \\ \textbf{Training}\end{tabular} & \textbf{Para.} & \textbf{PSNR-$\mu$} & \textbf{PSNR-L} \\ \midrule
{Baseline} & \ding{56} & \ding{56} & \ding{56} & 5.09M &  44.23    & 42.19     \\
{LF-Diff$_{s1}$} & \ding{52} & \ding{56} & \ding{56} & 5.24M & 50.42     & 50.89     \\
{Split training} & \ding{56} & \ding{52} & \ding{56} & 7.48M & 41.93     & 39.30     \\
{LF-Diff$_{s2}$ w/o DM} & \ding{56} & \ding{56} & \ding{52} & 5.85M & 44.45     &  42.35   \\
{LF-Diff$_{s2}$ w/ DM} & \ding{56} & \ding{52} & \ding{52} & 7.48M & 44.76   & 42.59  \\ \bottomrule
\end{tabular}}
\caption{Ablation studies of Diffusion Prior and Joint Training.}
\label{abdp}
\end{table}

\begin{table}[]\small
\centering
\scalebox{0.80}{
\begin{tabular}{c|ccccc}
\toprule
\textbf{Sampling step} & \textbf{PSNR-$\mu$}  & \textbf{PSNR-L}  & \textbf{SSIM-$\mu$}   & \textbf{SSIM-L}   & \textbf{Time(s)} \\
\midrule
\textbf{S=5}           & 39.76 & 41.33 & 0.9819 & 0.9878 & 0.049   \\
\textbf{S=10}          & 44.76 & 42.59  & 0.9919 & 0.9906 & 0.106   \\
\textbf{S=20}          & 44.73 & 42.54 & 0.9919 & 0.9906 & 0.208   \\
\textbf{S=50}          & 44.66 & 42.45 & 0.9919 & 0.9906 & 0.479   \\
\textbf{S=100}         & 44.65 & 42.44 & 0.9919 & 0.9906 & 0.971  \\
\bottomrule
\end{tabular}}
\caption{Ablation studies of various settings on the sampling step during the reverse process. 
Time(s) only denotes the time expenditure of the DM for the corresponding setting.}
\vspace{-0.2cm}
\label{absampletime}
\end{table}

\section{Conclusion}
The potential of diffusion models in HDR deghosting has shown promising results, particularly in achieving visually perceptible outcomes. Different from image synthesis which generates each pixel from scratch, HDR imaging provides several LDR images as reference. Thus, it is inefficient to reconstruct the full HDR image starting from pure Gaussian noise. In this paper, we propose an efficient diffusion model called LF-Diff, consisting of LPENet, DHRNet, and a denoising network. Specifically, we apply the DM in a compact latent space to predict low-frequency priors of HDR images. These prior features provide explicit guidance for the image reconstruction process, thereby enhancing the details of the reconstructed HDR images. Compared to traditional DM-based methods, LF-Diff achieves accurate estimations and reduces artifacts in reconstructed images with much lower computational cost.
\clearpage